\pdfoutput=1
\documentclass[11pt]{article}
\usepackage[final]{acl}
\usepackage{times}
\usepackage{latexsym}
\usepackage[T1]{fontenc}
\usepackage[utf8]{inputenc}
\usepackage{microtype}
\usepackage{inconsolata}

\usepackage{graphicx}
\usepackage{amssymb, bm}		
\usepackage{wrapfig}		
\usepackage{subfigure}		
\usepackage{color}			
\usepackage{xcolor}			
\usepackage{multicol}		
\usepackage{multirow} 		
\usepackage{bbding}
\usepackage[figuresright]{rotating}
\usepackage{verbatim}
\usepackage{siunitx}

\usepackage{times}
\usepackage{soul}
\usepackage{url}
\usepackage[utf8]{inputenc}
\usepackage{caption}
\usepackage{graphicx}
\usepackage{amsmath}
\usepackage{amsthm}
\usepackage{booktabs}
\usepackage{algorithm}
\usepackage{algorithmic}
\usepackage{tcolorbox}
\usepackage{hyperref}
\usepackage{url}
\newtheorem{theorem}{Theorem}[section]
\newtheorem{definition}{Definition}[section]

\usepackage{tabularx}
\usepackage{makecell}

\newcommand{\ie}{\textit{i}.\textit{e}.}

\title{Nash CoT: Multi-Path Inference with Preference Equilibrium}
\author{Ziqi Zhang$^{1}$\thanks{\,\, Equal contribution.},\,\,Cunxiang Wang$^1$\footnotemark[1],\,\,Xiong Xiao$^2$,\,\,Yue Zhang$^1$\thanks{\,\, Corresponding authors.},\,\,Donglin Wang$^1$\footnotemark[2]\\
 $^1$ School of Engineering, Westlake University\\
 $^2$ University of Cambridge, UK\\
 stevezhangz@163.com \\
 \{wangcunxiang, zhangyue, wangdonglin\}@westlake.edu.cn}
\begin{document}
\maketitle
\begin{abstract} 
Chain of thought (CoT) is a reasoning framework that can enhance the performance of Large Language Models (LLMs) on complex inference tasks. In particular, among various studies related to CoT, multi-path inference stands out as a simple yet effective improvement. However, there is no optimal setting for the number of inference paths. Therefore, we have to increase the number of inference paths to obtain better results, which in turn increases the inference cost. To address this limitation, we can utilize question-related role templates to guide LLMs into relevant roles, thereby increasing the possibility of correct inferences for each path and further reducing dependence on the number of inference paths while improving reasoning accuracy. However, placing LLMs into specific roles may reduce their reasoning diversity and performance on a few tasks where role dependence is low. To alleviate the excessive immersion of the LLM into a specific role, we propose Nash CoT by constructing a game system on each path that balances the generation from role-specific LLMs' and the general LLMs' generation, thereby ensuring both effective role adoption and diversity in LLM generation further maintaining the performance of multi-path inference while reducing the requirement of the number of inference paths. We evaluate Nash CoT across various inference tasks, including Arabic Reasoning, Commonsense Question Answering, and Symbolic Inference, achieving results that are comparable to or better than those of multi-path CoT with the equal number of inference paths.
\end{abstract}
\section{Introduction}
Large Language Models (LLMs) have profoundly revolutionized the field of Natural Language Processing (NLP)~\citep{ouyang2022training,touvron2023llama,jiang2023mistral,brown2020language,openai2024gpt4}. Specifically, leveraging human-designed instructions as input, LLMs demonstrate superior inference performance across various types of simple reasoning tasks~\citep{Radford2019LanguageMA, NEURIPS2020_1457c0d6}. 

However, in complex tasks, direct reasoning LLMs do not yield good results~\citep{rae2022scaling}. To improve LLMs' inference performance on complex inference tasks, it is popular to adopt a step-by-step reasoning framework known as Chain-of-Thought (CoT) prompting~\citep{wei2023chainofthought}. For example, when appending templates that can guide LLMs to perform step-by-step reasoning, such as \texttt{``Let's think step by step"}, to the end of given question. We can lead the LLMs to output question-related rationale and further obtain the answer, enabling LLMs to achieve better performance than zero-shot reasoning.
\begin{figure*}[ht]
\centering\hspace{-10pt}
\includegraphics[scale=0.65]{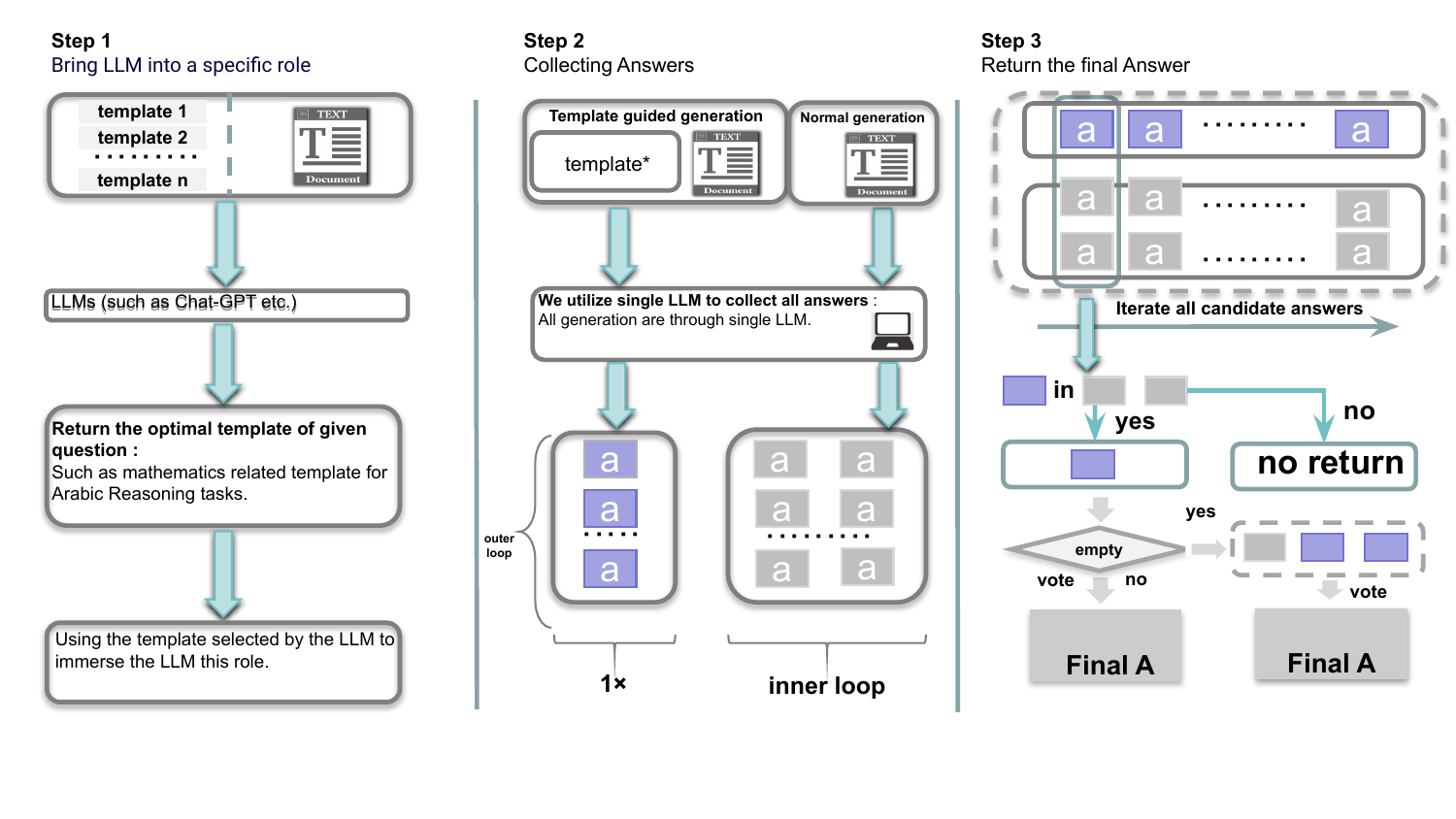}
\vspace{-30pt}
\caption{Demonstrations of Nash Chain-of-Thought (Nash CoT). As shown in this figure, Nash CoT can be divided into three main steps. \textbf{Step 1} involves bringing the LLM into a template-related role. \textbf{Step 2} utilizes the role-immersed LLM and LLM under normal conditions to collect model predictions separately. \textbf{Step 3} filters the responses to ensure the existence of a unique Nash Equilibrium (NE).}
\label{nash_cot}
\vspace{-2mm}
\end{figure*}
Subsequently, among various CoT-related improvements, modified by utilizing multi-path inference which is represented by~\citet{wang2023selfconsistency} is one of the most effective reasoning approaches. However, there is no theoretical proof identifying an optimal number of paths for multi-path reasoning. To ensure better inference performance, it is imperative to augment the inference paths. However, this might burden the inference budgets.

In order to reduce the dependency of multi-path inference on the number of inference paths while further enhancing its effectiveness, we can cast LLMs into a relevant role for a specific question. For instance, we can utilize a template such as "\texttt{You're a mathematician, and $\cdots$}" to cast the LLM in the role of a mathematician. This approach can improve the accuracy of each path's answer generated by the LLM, thereby reducing the multi-path CoT's reliance on the number of paths, potentially improving the LLM's reasoning.

However, excessively casting the LLM into a specific role using templates may compromise its robustness (diversity), thereby adversely affecting its performance in solving problems with minimal role dependency. To address this problem, we can introduce a certain degree of randomness into the reasoning process of the LLM. This can mitigate the constraints imposed by excessive role casting on the reasoning performance, ultimately maintaining the role-immersed multi-path CoT's reasoning performance in these questions. Driven by this motivation, we construct a game system on each path, where the LLM in the role-playing context and the LLM in its normal state act as two distinct players. When this game system formed by these two players reaches a unique Nash Equilibrium (NE), the reasoning under that path can balance the preferences of both role-playing with that of the LLM's normal state. Furthermore, we can choose the response with the most NE hits through voting to ensure the performance of LLM reasoning.
\begin{figure}[ht]
\centering
\includegraphics[scale=0.29]{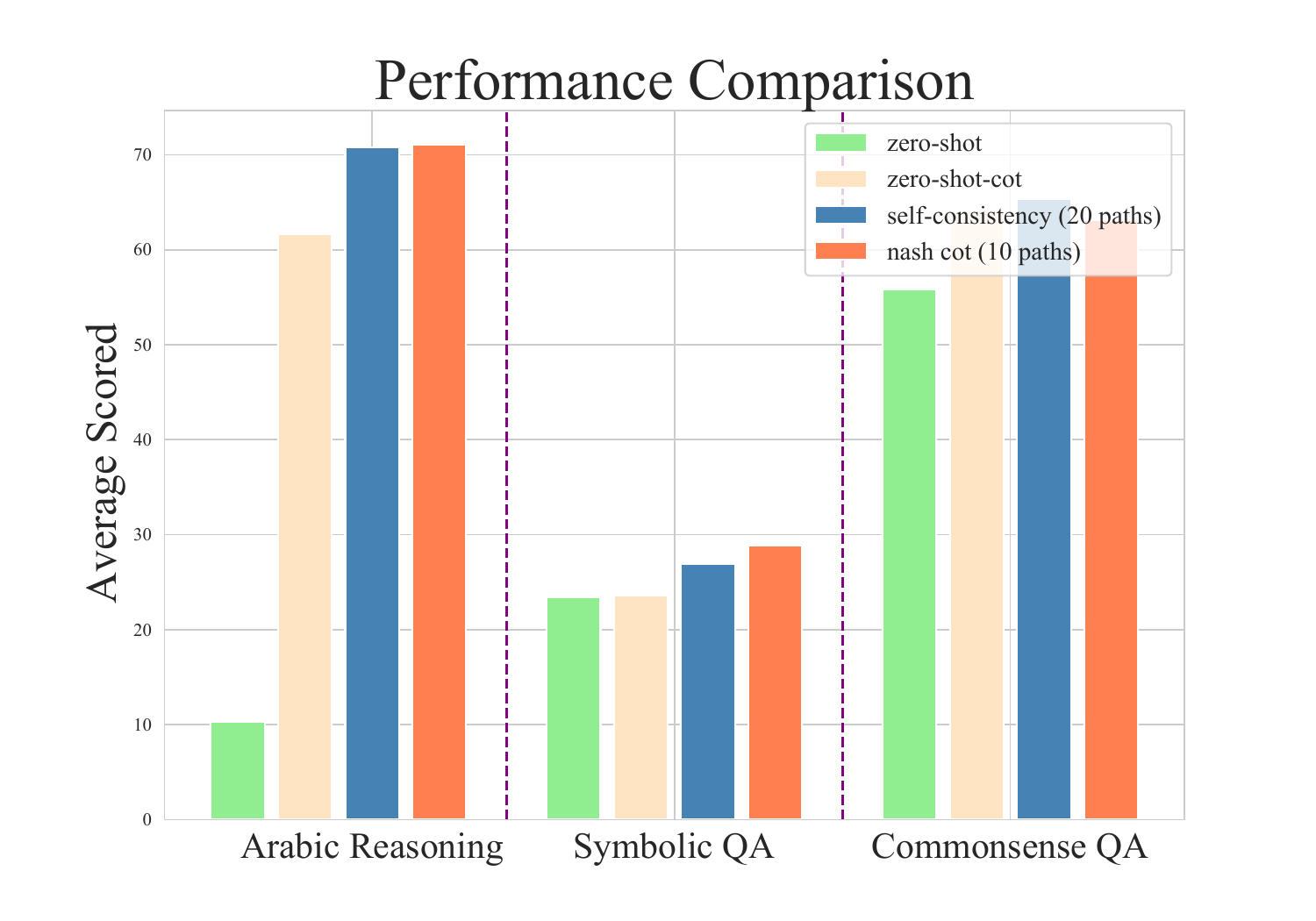}
\caption{General Performance Comparison. We compare the average performance of, zero-shot, and zero-shot CoT self-consistency (20 Paths) with our Nash CoT (10 Paths) on Mistral-Instruct and GLM4.}
\label{performance_comparsion}
\vspace{-2mm}
\end{figure}

We conduct experiments with two local deployed LLMs that include Mistral-Instruct \citep{jiang2023mistral} and GLM-4 \citep{zeng2022glm,du2022glm} on various inference tasks, including  Arabic Reasoning \citep{koncel-kedziorski-etal-2015-parsing,hosseini-etal-2014-learning}, Symbolic Reasoning \citep{wei2023chainofthought} and Commonsense Reasoning~\citep{talmor2019commonsenseqa,geva2021did}.
As shown in Figure~\ref{performance_comparsion},  Nash CoT can achieve similar or even better performance against self-consistency with fewer inference paths. 
Meanwhile, as shown in Figure~\ref{time_consuming}, Nash CoT can significantly reduce the inference cost by up to 50\% on locally deployed LLMs. To summarize, our contribution can be summarized as follows:
\begin{itemize}
    \item To our knowledge, we are the first to introduce the concept of NE into multi-path CoT, with the objective of balancing text generation guided by the role-immersed LLM with that produced by the LLM's default state. This innovation enables us to minimize the number of paths required to achieve good results in multi-path inference, while simultaneously preserving the multi-path CoT's inherent performance advantages.
    \item We evaluated the performance of Nash CoT on a wide range of reasoning tasks. Nash CoT can achieve similar results to self-consistency in Arabic Reasoning, Symbolic Reasoning, and Commonsense Question Answering tasks, given the same number of reasoning paths.\footnote{We have released our code at <\url{https://github.com/stevezhangzA/nash-chain-of-thought}>.} 
\end{itemize}

\section{Related Work}

\paragraph{Chain-of-Thought (CoT).} There are three majority types of CoT approaches. The first is zero-shot CoT, which involves prompting the LLM with simple instructions to guide its generation step by step~\citep{kojima2023large}. The second type is Manual CoT, which begins with sampling several cases from a dataset or manually designing templates. These cases are then used as demonstrations to guide the LLM in generating responses~\citep{wei2023chainofthought}. However, such methods can be biased if the demonstrations aren't sufficient to cover the optimal distribution. Automatic CoT~\citep{zhang2022automatic} first clusters the dataset, then samples the most representative sentence from each cluster as the demonstrations and finally uses these demonstrations to guide LLM inference. On the other hand, self-consistency~\citep{wang2023selfconsistency} showcases strong performance in vast benchmarks. Apart from its impact on inference performance, self-consistency also boasts scalability as a key advantage. It can seamlessly integrate with different approaches, such as tabular chain-of-thought (Tab-CoT) \cite{jin2023tabcot}, making it adaptable and versatile for various applications. Furthermore, despite self-consistency can improve LLM's performance on Arabic benchmarks, self-consistency must be inferred multiple times, burdening deployment budgets.

One way to overcome this limitation is to infer multiple paths and fine-tune them based on the most frequent path. Specifically,~\citet{huang2022large} proposes that by gathering inferences from multiple paths and sampling the most frequent generation, we can enhance the inference performance of a smaller LLM. However, this approach still requires updating the LLM's parameters, which is inefficient. Therefore, it is necessary to further develop inference methods to maintain the performance of self-consistency while reducing the number of multi-path CoT inference paths.

\paragraph{Preference Optimization.} Training policy with Reinforcement Learning (RL) to reflect preference, termed Reinforcement Learning with Human Feedback (RLFH), is initially introduced by~\citet{Akrour2011PreferenceBasedPL} and has since undergone consistent improvement and refinement by~\citet{Cheng2011PreferenceBasedPI,BusaFekete2013PreferencebasedED,Wilson2012ABA}. This approach has been widely applied to adjust the parameters of LLMs in order to align them with human preferences~\citep{ouyang2022training,jiang2023mistral,touvron2023llama}. Recently, a new approach called Direct Optimizing from Preference (DPO) has been proposed by~\citet{rafailov2023direct}, aiming to directly adjust LLMs to reflect human preferences without requiring a reward model. Additionally, \citet{munos2023nash} proposes combining DPO with NE to ensure the convergence of the last iterated policy. Despite our study also utilizing the concept of NE in the Preference model, the main difference is that we utilize NE as a standard to pick up the answer instead of optimizing the LLM's parameters.

\section{Preference Equilibrium\label{claim} in mini-batch inference} 
The experimental results of self-consistency indicate that multi-path reasoning can unleash the reasoning capability of LLMs. 
However, multi-path inference lacks a theoretical foundation for determining the optimal number of inference paths, requiring us to increase the number of paths to ensure performance, and further increase the computational resource consumption. To reduce the number of paths required for multi-path inference, we utilize the concept of NE to locally construct a binary game system in multi-path inference. Specifically, the preference of each valid inference path of the LLM needs to achieve NE with the preferences of the generation guided by the role-immersed LLM. This approach increases the possibility of each path correctly answering the question while maintaining a certain level of robustness (diversity), thereby reducing the number of inference paths required by self-consistency.

\paragraph{Preference Model.} Given text input $x$, and the sampled predictions (answers) $y_1$, $y_2$, and Bradley-Terry reward model $r_{\theta}$ that has been deliberated by~\citeauthor{chen2024selfplayfinetuningconvertsweak} We first define $y_1$ is preferred over $y_2$ as Equation~\ref{preference}: 
\begin{align}
\begin{split}
\mathcal{P}(y_2\prec y_1|x):={\sigma}(r_{\theta}(y_1|x)-r_{\theta}(y_2|x)),
\end{split}
\label{preference}
\end{align}
where $\mathcal{P}$ is preference model reflects the preference when inputting the pairs $(y_1,y_2)$. 

\begin{tcolorbox}[title=\textbf{\textbf{Player Templates} : Role templates for our LLM that are shown the structure: \{\texttt{id(player): description}.\}}]

\textbf{Mathematician:} You are a mathematician, you excel at analyzing problems from a mathematical logical perspective and arrive at conclusions that align with your values.\\
\noindent
\textbf{Literary scholar:} You are a literary scholar who has read a vast array of literary works. Please consider the problem from the perspective of a literary scholar.\\
\noindent
\textbf{Philosophical:} You are a philosopher, your knowledge base includes a wealth of philosophical knowledge. You enjoy approaching problems from a philosophical perspective and arriving at conclusions that align with your values.\\
$\cdots$ \texttt{(other cases have been appended to the Appendix.)}
\label{demonstration}
\end{tcolorbox}

Furthermore, drawing from the definition in~\citep{munos2023nash}, we define one policy (LLM for decision making) $\pi_1$ as more preferred over another policy $\pi_2$ as: 
\begin{align}
\begin{split}
    \mathcal{P}(\pi_2\prec \pi_1):=\mathbb{E}_{^{y_1\sim \pi_{1}(\cdot|x)}_{y_2\sim\pi_{2}(\cdot|x)}}\bigg[\mathcal{P}(y_2\prec y_1|x)\bigg],
\end{split}
\end{align}
subsequently, we imply the necessity and existence of a unique NE in the Preference model. 
\paragraph{Preference Equilibrium.} We first discuss why we should construct a game system: To increase the accuracy of each inference path, we can select the most suitable template (example~\ref{demonstration}) for the given question, immersing the LLM into a template-related role. This approach helps LLMs to solve problems more effectively, thereby reducing the number of inference paths needed.

However, this may lead to some issues: If the LLM is excessively immersed in a specific role, it may reduce the diversity of LLM's generation, and further reduce LLM's accuracy in solving tasks that have low role dependence. To address these problems, we locally build a \textbf{bi-player game system} that the preference of role-immersed LLM (\texttt{player 1}) over the normal status of LLM (\texttt{player 2}) is the pay-off of role-immersed LLM, vice visa. If \texttt{player 1} and \texttt{player 2} reach NE, then the generation can balance the preference of both \texttt{player 1} and \texttt{player 2}.

\paragraph{Regularized form of preference model.}
Furthermore, we study a preference model with a Kullback-Leibler (KL) constraint $\ie$
\begin{align}
\small
\nonumber
\begin{split}
\mathcal{P}_{\tau}(\pi_1\prec\pi_2)\overset{\texttt{def}}{=}\mathcal{P}(\pi_1\prec\pi_2)-\tau\cdot {\rm KL}(\pi_2(\cdot|x)||\mu(\cdot|x))+\\\tau\cdot {\rm KL}(\pi_1(\cdot|x)||\mu(\cdot|x)),
\end{split}
\end{align} Unlike the trainable RLHF framework proposed by~\citeauthor{munos2023nash}, which aims to regularize the training policy to be close to a known reference (safe) policy $\mu$, our inference system is not trainable. Instead, the incorporation of this regularization term serves a different purpose: \textit{Facilitate a theoretical analysis that demonstrates that, under the KL constraint, the game system possesses a unique Nash Equilibrium (NE).}
\begin{definition}[Preference Equilibrium] Given any two policy (players) $\pi_1, \pi_2$ along with the pay-off of $\pi_1, \pi_2$ as $\mathcal{P}(\pi_2\prec \pi_1)-\tau\cdot{\rm KL}(\pi_1||\mu)$ and $\mathcal{P}(\pi_1\prec \pi_2)-\tau\cdot{\rm KL}(\pi_2||\mu)$, we define the NE of $\pi_1$, $\pi_2$ as Preference Equilibrium. 
\label{def1}
\end{definition}
Subsequently, we define the status that \texttt{player 1} and \texttt{player 2} reach NE as \textbf{Preference Equilibrium} (\textbf{Definition~\ref{def1}}). Meanwhile, in Theorem~\ref{lemma1}, we prove the uniqueness of NE in this system. Specifically, having the strategy of \texttt{player 1} equal to \texttt{player 2} is a crucial prerequisite for ensuring that the game system has a unique \textbf{Preference Equilibrium}.
\begin{theorem}[Uniqueness of Preference Equilibrium] Given any two policy (player) $\pi_1$ and $\pi_2$ within the game system defined in Definition 1, where $\pi \in \Pi$. $\pi_1=\pi_2$ is an  essential condition to guarantee this system has a unique NE.
\label{lemma1}
\end{theorem}
\noindent
\textbf{\textit{Proof}} of Theorem~\ref{lemma1} see Appendix~\ref{proof}.
\paragraph{Meanings of the existence of Preference Equilibrium.} When reaching Preference Equilibrium, the preference of generations made by the role-immersed LLM is aligned with those made by the LLM without templates. Specifically, the preference for role-immersed LLM generation is much closer to the selected role-immersed context, while those of the normal status LLM are predominantly based on its parameters which are much more diverse than role-immersed LLM generation. Therefore, Preference Equilibrium can balance the requirement of role-immersed LLM generation and the diversity of the LLM generation in each path. 
\paragraph{Meanings of $\pi_1=\pi_2$.} 
Theorem~\ref{lemma1} proves $\pi_1=\pi_2$ is an essential condition that guaranteeing this game system has a unique NE. Therefore, we can choose $\pi_1=\pi_2$ as the indicator to ensure that this game system has a unique NE.
\paragraph{Meanings of KL terms.} If the payoff of $\pi$  lacks a KL term, proving the uniqueness of the Nash Equilibrium (NE) becomes infeasible. Furthermore, the KL term plays a crucial preferential role in penalizing role-immersed generation, particularly when it diverges significantly from normal generation. This penalty helps to strike a balance between role-immersed and normal generations (see the \textbf{\textit{Proof}} of Theorem~\ref{lemma1}).
\subsection{Mini-batch inference with Preference Equilibrium\label{conception}} Subsequently, based on Preference Equilibrium, we conceptualize a mini-batch inference (shown in \textbf{Step 2} of Figure~\ref{nash_cot}) as a bi-player game system. This approach aims to maintain the performance benefits of multi-path inference while retaining some of the inherent randomness (diversity) of standard inference methods. Before introducing mini-batch inference, we first define $x^{t}$ as the template of zero-shot CoT. Meanwhile, we leverage LLMs to select a player template from the candidate set $\{x^{c}_0,x^{c}_1,\cdots,x^{c}_n\}$ for role-immersed generation, guided by Equation~\ref{preference}. 
\begin{algorithm}[ht]
\small
\caption{~Nash CoT (Answer Gathering)}
\hspace{1pt} \textbf{Require:} Candidate question q sampled from $\mathcal{Q}=\{q_0,q_2,\cdots,q_n\}$; Outer iterations ${\rm n}_{\rm outer}$; Num of mini-batch inference $\rm n_{mini}$; Large language model $\pi$; CoT prompt $x^{t}$, candidiate player template $\{x^{c}_0,x^{c}_1,\cdots,x^{c}_n\}$, ${\rm Prompt}(\{x^{c}\})$ is used to point out the most preferred $x^{c}$.\\
\hspace{1pt} \textbf{Generation:}
\begin{algorithmic} [1]
\STATE Initialize answer list ${\rm ans}=[\,]$.
\STATE $x_{c}\leftarrow \pi(\cdot|{\rm Prompt}(\{x^{c}\}))$
\FOR{t in range(${\rm n}_{\rm outer}$)}
\STATE Initialize prefer pairs ${\rm pref}=[\,]$.
\FOR{t in range($\rm n_{mini}$)}
\STATE $y \leftarrow \pi(\cdot|x^{t},q)$;
\STATE ${\rm ans}.$append($y$);
\ENDFOR
\STATE $y^*\leftarrow \pi(\cdot|x^{c},x^{t},x)$ 
\STATE $\tau.{\rm append}([y^*,{\rm ans}])$
\ENDFOR
\STATE Return $\tau$
\end{algorithmic}
\label{stage1}
\end{algorithm}
In terms of the process of mini-batch inference, we firstly infer LLM twice times $\ie$ $[y_0,y_1]\leftarrow [\pi(\cdot|x^{t},x),\pi(\cdot|x^{t},x)]$, in particular, we have conducted ablations about the setting of 'twice' in section ablation. Meanwhile, due to the inherent uncertainty (diversity) of LLM, the generation of $[y_0, y_1]$ can be considered a potential set of distinct predictions. Subsequently, the role-immersed generation can be sampled by querying LLM with $x^{c}$ and $x^{t}$ $\ie$ $y^*\leftarrow \pi(\cdot|x^{c},x^{t}, x)$. Furthermore, we can select an answer from $y_1$ and $y_2$ that is the same as $y^*$ to guarantee the uniqueness of NA. 

Subsequently, we further introduce Nash CoT in the next chapter. (\textit{In particular, in the context of current chapter, successful inference cannot always be guaranteed. $\ie$, $y^*$ may not always in $[y_1,y_2]$. We address this problem in the following chapters.})
\section{Nash Chain of Thought (Nash CoT)}
Nash CoT can be regarded as an extension of mini-batch inference with Preference Equilibrium, implementing multiple mini-batch inferences to further enhance performance. This approach is inspired by self-consistency experiments, which suggest that increasing the number of paths enhances LLM's inference accuracy. Specifically, Nash CoT's reasoning process for each question can be divided into two stages: Answer Gathering and Answer Filtering.
\paragraph{Answer Gathering.} When generating candidate answers, the process predominantly involves two types of loops: \textbf{mini-batch Loops ($\rm n_{\rm mini}$)}: As shown in Algorithm~\ref{stage1}, this process involves searching for role-immersed generations within two rounds of generation $[y_1,y_2]$. We denote the times of these two predictions as the $\rm n_{\rm mini}$. Moreover, to mitigate the impact brought by low-frequency predictions, we introduce iterating $\rm n_{\rm mini}$ multiple times. This leads us to another type of loop: \textbf{Outer Loops ($\rm n_{\rm outer}$)}: This loop resembles the concept of multi-path inference that iterates $\rm n_{\rm mini}$ multiple times. After completing loop $\rm n_{\rm outer}$, We filter the generated answers and retain the one that satisfies the uniqueness of the NE most frequently. (shown in Algorithm~\ref{stage2}), as the predicted answer.
\begin{algorithm}[ht]\small
\caption{~Nash CoT (Answer Filtering)}
\hspace{1pt} \textbf{Require:} Preference pair list $\tau$ \\
\hspace{1pt} \textbf{Filtering:}
\begin{algorithmic} [1]
\STATE Initialize hash table: ${\rm hash}=\{\}:k\rightarrow v$.\\
\STATE Initialize new answer list ${\rm n\-ans}=[\,]$.
\FOR{$[y^*,{\rm ans}]_i$ in $\tau$}
\IF{$y^*\in {\rm ans}$}
\STATE ${\rm hash}$[$y^*$]+=1\\
\ENDIF \\
${\rm n\-ans}.{\rm extend}([y^*,{\rm ans}[0],{\rm ans}[1]])$
\ENDFOR
\IF{${\rm hash}\equiv \{\}$}
\RETURN the most frequent y in ${\rm n\-ans}$
\ELSE
\RETURN $y\leftarrow k=\arg\max_{v} {\rm hash}$ 
\ENDIF
\end{algorithmic}
\label{stage2}
\end{algorithm}
\begin{table*}[ht]
\centering
\resizebox{\linewidth}{!}{
\begin{tabular}{ccccccccc}
\toprule
\textbf{Core LLM}&\textbf{Methods}&\textbf{SingleEQ}&\textbf{AddSub}&\textbf{MultiArith}&\textbf{GSM8K}&\textbf{AQuA}&\textbf{SVAMP}&\textbf{Avg.}\\
\midrule
\multirow{4}[0]{*}{Mistral-Instruct (7B) }&zero-shot&15.3$\pm$0.8	&12.0$\pm$2.8&3.3$\pm$1.3&2.7$\pm$2.0&20.8$\pm$1.5&7.7$\pm$2.0&10.3$\pm$7\\
&zero-shot CoT&76.0$\pm$0.8&82.5$\pm$2.0&75.4$\pm$6.1	&44.3$\pm$ 4.0	&27.9$\pm$2.3&63.4$\pm$6.9&61.6$\pm$19\\
&self-consistency (20 Paths)&82.5$\pm$0.8&86.3$\pm$5.1&86.3$\pm$2.8&58.5$\pm$2.8&34.4$\pm$6.1	&76.5$\pm$2.8&70.8$\pm$19\\
&Nash CoT (10 Paths)&81.4$\pm$0.8&86.3$\pm$6.0&86.3$\pm$4.7&55.7$\pm$5.8&39.9$\pm$5.4	&77.0$\pm$3.5&71.1$\pm$17\\
\hline
\multirow{4}[0]{*}{GLM4-chat (9B)}&zero-shot&1.1$\pm$1.5&1.1$\pm$1.5&12.6$\pm$3.9&12.0$\pm$2.0&22.4$\pm$4.1&4.4$\pm$2.8&8.9$\pm$8 \\
&zero-shot CoT&90.7$\pm$1.5&90.7$\pm$1.5&98.4$\pm$1.3&80.9$\pm$2.8&20.8$\pm$ 3.1	&86.9$\pm$3.5&78.1$\pm$26 \\
&self-consistency (20 Paths)&92.3$\pm$2.0&90.2$\pm$2.3&98.4$\pm$2.3&89.3$\pm$0.2&20.8$\pm$3.1	&91.5$\pm$1.1&80.4$\pm$27 \\
&Nash CoT (10 Paths)&91.3$\pm$0.8&90.2$\pm$2.7&96.7$\pm$3.3&80.3$\pm$1.3	&20.8$\pm$3.1&88.0$\pm$2.0&77.9$\pm$26 \\
\bottomrule
\end{tabular}}
\caption{Experimental results on Arabic Reasoning benchmarks. We test Zero-Shot CoT and Nash CoT with the core LLM includes Mistral-Instruct (7B) and GLM4-chat (9B) on mathematical benchmarks including AddSub, MultiArith, SingleEQ, SVAMP, GSM8K, and AQuA. Nash CoT performs the best.}
\label{tab:main_label1}
\end{table*}
\paragraph{Answer Filtering.} In terms of answer filtering, as shown in Algorithm~\ref{stage2} we first count the most frequent prediction to satisfy the uniqueness of the NE. Specifically, we count all $y^*$ satisfy $y^*\in [y_1,y_2]$ and compute their frequency. Subsequently, we return the most frequent answer. Otherwise, if no cases satisfy $y^*\in [y_1,y_2]$, we adapt answer filtering by selecting the most frequent prediction among all generated answers.
\begin{tcolorbox}[title=\textbf{Preference Templates : utilized to confine the prompt for preference model.}]
\textbf{Q:} Current issue is $\textbf{\{query\}}$, and the best player is who? Please give us the number of that player from the options below: $\textbf{\{description\}}$. There are total $\textbf{N(\{key(player)\})}$ players including $\textbf{\{key(player)\}}$. Please point out the most appropriate player for the following task: $\textbf{candidate questions}$\\
\textcolor{gray}{ \textbf{A:} Let us think step by step. $\rightarrow$ \texttt{z} {(obtain the rational z)}}\\
\textbf{A:} Let us think step by step. + \texttt{z}+ Therefore, the most appropriate player in this game is who? (please direct give us the number)
\label{demonstration2}
\end{tcolorbox}
\paragraph{Practical Implementation of Preference model $\mathcal{P}$.} In the process of practical implementation, we do not explicitly train a reward model $r_{\theta}$ to confine the player template $x^{c}$ utilizing Equation~\ref{preference}. Instead, we directly utilize the template (shown in \textbf{Preference Template}) to guide the LLM in determining the most suitable player template for a given question. For example, when presented with a coin flip question as shown in Figure~\ref{demonstration2}, we fill the \textbf{Preference Template} with the given question and \textbf{Player Templates}. This filled template is then input into the LLM to provide the id of the most suitable player template from the available options. In particular, we consider utilizing LLM to search templates to be effective, this is because most of the baselines we selected have been turned to reflect human preference, thus LLM can be directly utilized as the preference model to point out the most preferred option among candidate options (we provide cases in section~\ref{appendix_data}). 
Subsequently, we propose Nash CoT, which iterates through Algorithm~\ref{stage1} and Algorithm~\ref{stage2} to perform inference on all sampled questions, where $\tau$ represents the candidate answers from the Answer Gathering stage.
\section{Experiments}
\begin{table*}[ht]
\small
\centering
\resizebox{\linewidth}{!}{\begin{tabular}{ccccccccc}
\toprule
\textbf{Core LLM}&\textbf{Methods}&\textbf{Coin-Flipping}&\textbf{Last Letters}&\textbf{Object Tracking}&\textbf{Bigbench Date}&\textbf{Avg.}\\
\midrule
\multirow{4}[0]{*}{Mistral-Instruct (7B) }&zero-shot&26.8$\pm$5.1	&0.0$\pm$0.0&35.5$\pm$4.1&31.1$\pm$7.6&23.4$\pm$14\\
&zero-shot CoT&27.9$\pm$4.0&0.0$\pm$0.0&30.1$\pm$2.8&36.6$\pm$5.4&23.6$\pm$14 \\
&self-consistency (20 Paths)&21.9$\pm$4.7&0.0$\pm$0.0&38.8$\pm$0.8&47.0$\pm$1.5&26.9$\pm$18 \\
&Nash CoT (10 Paths) &29.0$\pm$5.4&0.5$\pm$0.08&44.8$\pm$2.0&41.1$\pm$1.2&28.9$\pm$17 \\
\hline
\multirow{4}[0]{*}{GLM4-chat (9B)}&zero-shot&27.3$\pm$6.3&0.0$\pm$0.0&38.8$\pm$0.8&16.4$\pm$2.3&20.7$\pm$14 \\
&zero-shot CoT&87.4$\pm$0.8&0.0$\pm$0.0&37.7$\pm$2.3&16.4$\pm$4.8&35.4$\pm$33 \\
&self-consistency (20 Paths)&98.9$\pm$1.5&0.0$\pm$0.0&37.7$\pm$2.3&16.4$\pm$4.8&38.3$\pm$38 \\
&Nash CoT (10 Paths)&93.4$\pm$2.7&0.0$\pm$0.0&37.7$\pm$2.3&16.4$\pm$4.8&36.9$\pm$35 \\
\bottomrule
\end{tabular}}
\caption{Experimental results on symbolic inference benchmarks. We test Zero-Shot CoT and Nash CoT with Mistral-Instruct (7B) and GLM4-chat (9B) on Symbolic QA benchmarks including Coin-Flipping, Last Letters, and Object Tracking. Among these baselines, Nash CoT performs the best.}
\label{tab:main_label2}
\end{table*}
\begin{table*}[ht]
\centering
\small
\begin{tabular}{ccccccccc}
\toprule
\textbf{Core LLM}&\textbf{Methods}&\textbf{StrategyQA}&\textbf{CommonsensQA}&\textbf{Avg.}\\
\midrule
\multirow{4}[0]{*}{Mistral-Instruct (7B)}&zero-shot&49.2$\pm$8.8&62.3$\pm$4.8&55.8$\pm$7 \\
&zero-shot CoT&57.4$\pm$2.3&70.5$\pm$2.7&64.0$\pm$7 \\
&self-consistency (20 Paths)&59.6$\pm$2.0&71.0$\pm$3.4&65.3$\pm$6 \\
&Nash CoT (10 Paths)&56.8$\pm$2.0&69.4$\pm$4.7&63.1$\pm$6\\
\hline
\multirow{4}[0]{*}{GLM4-chat (9B)}&zero-shot&56.8$\pm$4.7&17.5$\pm$2.0&37.2$\pm$20 \\
&zero-shot CoT&63.9$\pm$2.3&18.0$\pm$2.3&41.0$\pm$23\\
&self-consistency (20 Paths)&69.9$\pm$3.3&18.0$\pm$2.3&44.0$\pm$26 \\
&Nash CoT (10 Paths)&66.7$\pm$0.8&18.0$\pm$2.3&42.4$\pm$24 \\
\bottomrule
\end{tabular}
\caption{Experimental results on Commonsense Reasoning. We test Zero-Shot CoT and Nash CoT with Mistral-Instruct (7B) and GLM4-chat (9B) on Commonsense Reasoning datasets includes StrategyQA and CommonsenseQA .}
\label{tab:my_labe3}
\vspace{-6mm}
\end{table*}
The goal of our experiment is to 1) showcase the performance advantage and effectiveness of Nash CoT. 2) showcase whether Nash CoT helps reduce inference paths and inference time. 
\paragraph{Datasets.} Our majority benchmarks are composed of three different kinds of inference tasks. 1) \textit{Arabic Reasoning:} SingleEq~\citep{koncel-kedziorski-etal-2015-parsing}, AddSub~\citep{hosseini-etal-2014-learning}, MultiArith~\citep{roy2016solving}, GSM8K~\citep{cobbe2021training}, AQUA~\citep{ling2017program}, and SVAMP~\citep{patel-etal-2021-nlp}. 2) \textit{Symbolic Reasoning:} Last Letters, Coin Flip~\citep{wei2023chainofthought}, and Object Tracking, Bigbench Date. 3) \textit{Commonsense Question Answering:} CommonsenseQA~\cite{talmor2019commonsenseqa} and StrategyQA~\cite{geva2021did}. For more details about the dataset please refer to Appendix~\ref{appendix_data}.

\paragraph{LLMs.} To validate that Nash CoT is a general CoT method, we selected different large models as test models, including Mistral-Instruct (7B)~\citep{jiang2023mistral}, Chat-GLM4 (9B)~\cite{zeng2022glm,du2022glm}. In particular, all of these selected LLMs are turned via RLHF, and the difference between LLM turned with RLHF and the original foundation models have been detailed by~\citeauthor{ouyang2022training}

\paragraph{Baselines.} The preliminary baselines we utilized include zero-shot, zero-shot CoT~\citep{wei2023chainofthought}, and self-consistency~\citep{wang2023selfconsistency}. We test these approaches with frozen LLMs.  

\paragraph{Settings.} Our evaluation of all selected tasks utilizes the same experimental settings below:
\begin{itemize}
    \item \textbf{zero-shot and zero-shot CoT.} We follow the method proposed by \citet{wei2023chainofthought} and use the original template (e.g., \texttt{"Let's think step by step"}) for evaluation.
    \item \textbf{self-consistency.} We follow~\citeauthor {wang2023selfconsistency} to evaluate the performance of self-consistency with selected LLMs, utilizing the zero-shot CoT template. Additionally, we set the number of inference paths to 20.
    \item  \textbf{Nash CoT.} We set up $n_{\rm outer}$ as 3 and $n_{\rm mini}$ as 2, resulting in a total of $n_{\rm outer} \times (n_{\rm mini} + 1) + 1 = 10$ paths. Additionally, we have provided the \textbf{Player Templates} $x^{t}$ in Table 1 and the Appendix, meanwhile, we utilizing the same CoT template $x^{c}$ as in zero-shot CoT.
\end{itemize}
In particular, in Table~\ref{tab:main_label1},~\ref{tab:main_label2} and~\ref{tab:my_labe3}, we set the total number of reasoning paths for Nash CoT to 10 and for self-consistency to 20. The primary goal is to demonstrate that Nash CoT can achieve the same or even better results with only half the number of reasoning paths compared to self-consistency.

Additionally, all evaluations are conducted on the inference of 60 random sampled questions per seed. We have provided the mean and standard error in the majority tables.
\begin{figure}[ht]
    \centering
    \includegraphics[scale=0.43]{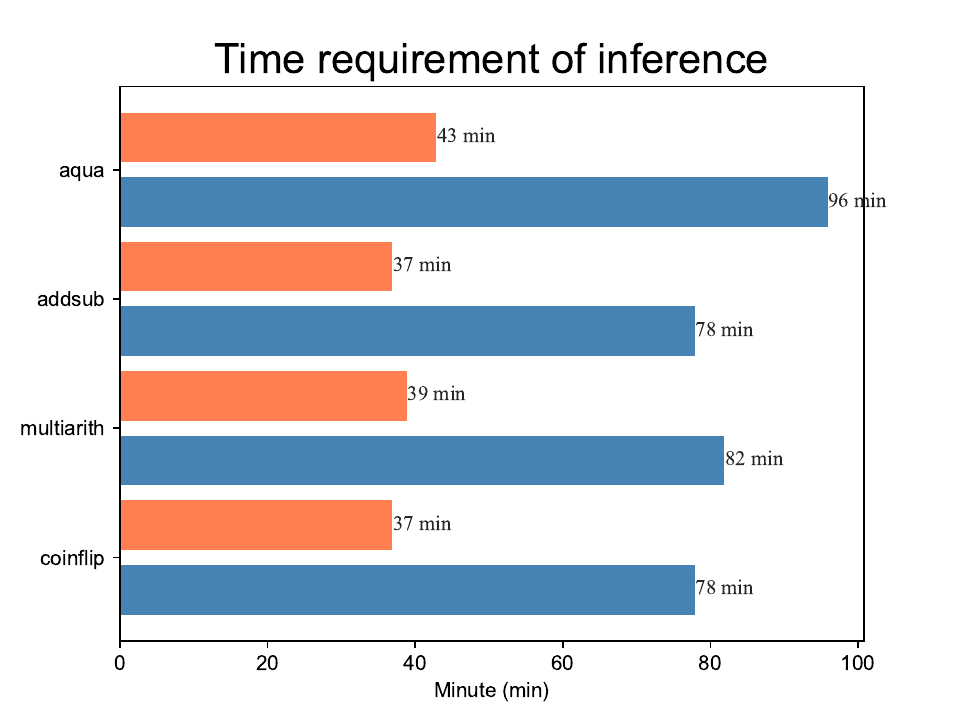}
    \vspace{-5pt}
    \caption{We used GLM4-chat (9B) on the same type of GPU (A100) to evaluate Nash CoT and self-consistency across selected tasks (60 questions per task). Nash CoT, employing a total of 10 paths, requires nearly half the time of self-consistency, which has 20 paths in total.}
    \label{time_consuming}
    \vspace{-2mm}
\end{figure}
\begin{figure*}[ht]
\centering
\includegraphics[scale=0.4]{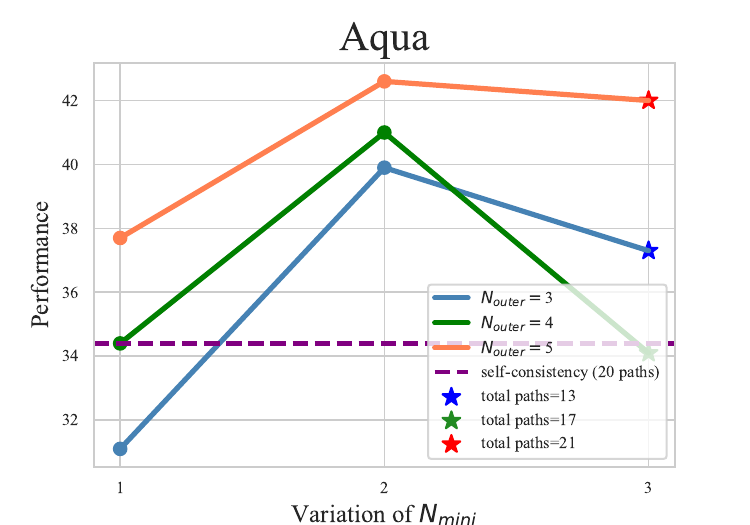}
\includegraphics[scale=0.4]{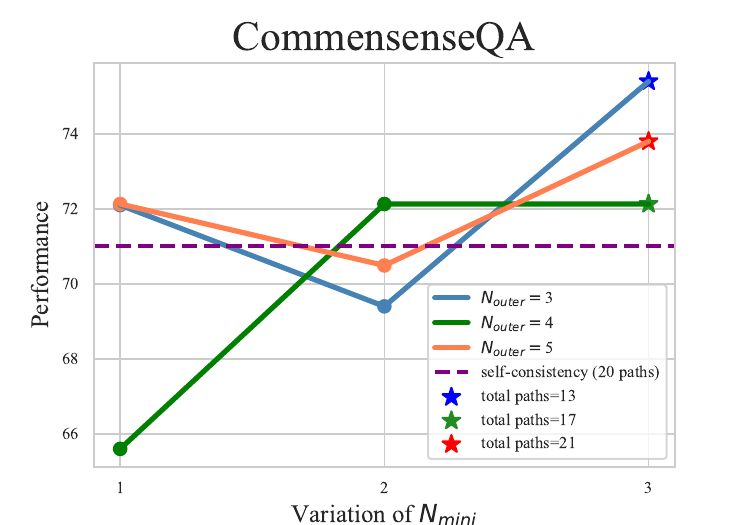}
\includegraphics[scale=0.4]{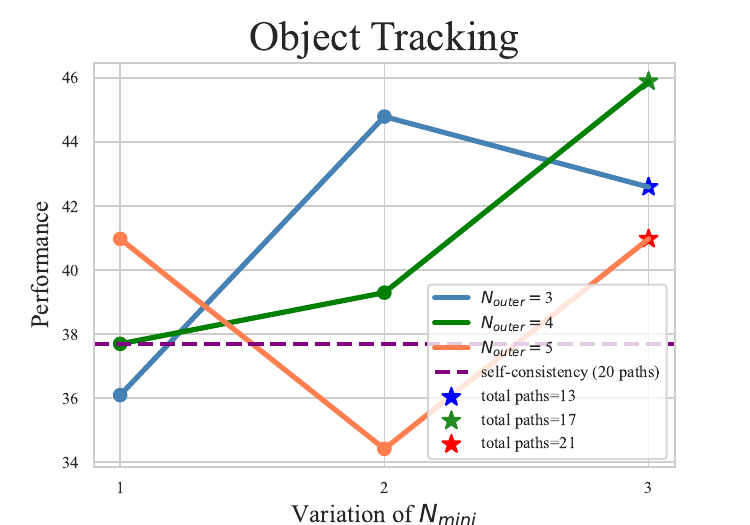}
\caption{We use Mistral-Instruct (7B) to examine the impact of loop numbers on the inference performance of the large language model. Specifically, we used solid lines of specific colors to represent the experimental performance under certain $N_{\rm outer}$ as the $N_{\rm mini}$ changed. We marked self-consistency with 20 paths using dashed lines, and some results of Nash CoT, with total paths close to 20, were marked with stars. }
\label{variation_loops}
\end{figure*}
\subsection{Experimental Results}
\begin{table*}[ht]
\centering
\resizebox{\linewidth}{!}{
\begin{tabular}{ccccccccc}
\toprule
Models&\textbf{Settings}&\textbf{Cnt and Prob}&\textbf{Precalculus}&\textbf{Inter-Alg}&\textbf{Number Theory}&\textbf{Avg.} \\
\midrule
\multirow{2}[0]{*}{Mistral-Instruct (7B)}&zero-shot-cot&40.9&16.4&16.4&24.6&24.6\\
&Nash CoT &64.8&\textbf{38.5}&36.9&\textbf{27.9}&\textbf{42.0} \\
&self-consistency&\textbf{72.1}&34.4&34.4&21.3&40.6 \\
\midrule
\multirow{2}[0]{*}{GLM4 (large)}&zero-shot-cot&83.6&90.2&85.2&91.9&87.7 \\
&Nash CoT&\textbf{95.1}&90.2&\textbf{98.3}&98.3&\textbf{95.5} \\
&self-consistency&93.3&90.2&95.1&100.0&94.7 \\
\bottomrule
\end{tabular}}
\caption{Equal Path Evaluation. This experiment is conducted to verify the performance of Nash CoT when the number of inference paths is set equal to self-consistency. In particular, Cnt and Prob denote Counting and Probability, while Inter-Alg denotes Intermediate Algebra.}
\label{larger_llm_performance}
\end{table*}
\paragraph{Evaluated Scores.} The majority experimental results are demonstrated in Table~\ref{tab:main_label1},~\ref{tab:main_label2} and~\ref{tab:my_labe3}. Nash CoT can improve Mistral-Instruct (7B) on almost all selected inference tasks while showcasing similar performance to self-consistency with twice inference paths on GLM4-chat (9B). In particular, we have provided the total paths of Nash CoT that it only require half of self-consistency, thus our claim in section~\ref{claim} can be validated.
When focusing on Mistral-Instruct (7B), Nash CoT has better performance on Arabic and Symbolic inference tasks, showcasing its superior performance on logic/math inference tasks. However, Nash CoT does not showcase improved performance in Commonsense Question Answering tasks. We argue that this is because Commonsense Question Answering tasks are more diverse, and the player template can't cover all topics. Therefore, the predefined player template limits Nash CoT on Commonsense Question Answering tasks. Importantly, we evaluate Nash CoT by utilizing only a total of 10 paths for inference in this section. However, additional experimental results in the ablation section show that Nash CoT outperforms self-consistency by increasing the inference loops. Meanwhile, we also evaluate by setting equal paths on much more challenging MATH~\cite{hendrycksmath2021}, showcasing that Nash CoT can outperform self-consistency. 

\paragraph{Comparing Nash CoT with self-consistency with equal paths.} To further evaluate the performance of Nash CoT, we utilize Mistral-Instruct (7B) while selecting one of state-of-the-art (SOTA) LLMs named GLM4 (large)~\citep{glm2024chatglm}, for comparison using an equal number of reasoning paths (10 paths). As shown in Table~\ref{larger_llm_performance}, Nash CoT outperforms self-consistency, which demonstrates the superiority of Nash CoT compared to self-consistency when the same number of path settings are set.

\paragraph{Inference Time.} The path of Nash CoT is composed of three different kinds of types $\ie$ zero-shot CoT for problem inference (in loop $N_{\rm mini}$), zero-shot CoT for player confining (in the outside loop of $N_{\rm outer}$), and player role-immersed CoT inference. Accordingly, the different path requires different inference time. Therefore, we further count the total time requirement of self-consistency and Nash CoT in Figure~\ref{time_consuming}, Nash CoT requires fewer inference time.

\section{Ablation Study} 
We further conduct extensive ablations to answer the following questions: 1) Is there a correlation between the performance of Nash CoT and the setting of the number of inference paths and can the performance of Nash CoT be improved to surpass self-consistency by adjusting the number of inference paths? 2) Does the template improve the accuracy of path predictions, and what impact does it have on experimental performance? 3) Is the NE the reason for the performance improvement observed in Nash CoT?

\paragraph{As the number of inference paths increases, Nash CoT can obviously surpass self-consistency with fewer inference paths.} To address question 1), we selected Mistral-Instruct (7B) and evaluated three different reasoning tasks, adjusting the $N_{\rm mini}$ and $N_{\rm outer}$. As shown in Figure~\ref{variation_loops}, as the number of loops increases, Nash CoT has a high possibility of outperforming self-consistency with fewer paths. 
However, different from self-consistency, the experimental results of Nash CoT do not show a monotonic (linear) relationship with the total number of paths.  This indicates that there is a significant difference between Nash CoT and self-consistency. Unlike Nash CoT, the experimental results of self-consistency show a clear improvement in performance as the number of paths increases.

\paragraph{The performance is impacted by the player template.} To illustrate the impact of the template, we removed the mathematical templates from the \textbf{Player Templates} and then evaluated Nash CoT on selected Arabic reasoning. Results are shown in Table \ref{tab:performace_decreasing}, showing an approximately 9.2\% decrement in \textbf{GSM8K} and 6.2\% decrement in \textbf{SVAMP}. Therefore, the performance of Nash CoT is impacted by the \textbf{Player Templates}.
\begin{table}[ht]
\small
\centering
\begin{tabular}{ccc}
\toprule
\textbf{GSM8K}&\textbf{AQuA}&\textbf{SVAMP} \\
\midrule
55.7$\rightarrow$ 50.6& 39.9$\rightarrow$ 39.8 & 77.0 $\rightarrow$72.2\\
\bottomrule
\end{tabular}
\caption{Performance decreasing. We remove the mathematics from \textbf{Player Templates} and test Nash CoT on selected Arabic Reasoning tasks.}
\label{tab:performace_decreasing}
\end{table}
As shown in Table~\ref{role_immersed}, we used three role templates with varying degrees of relevance to mathematical problems: mathematician, student, and poker player, to test the performance of Mistral-Instruct (7B) on GSM8K. Role templates with higher relevance to mathematical problems can bring more improvements to Mistral, further confirming the rationality of enhancing reasoning performance by substituting LLM into roles using templates.
\begin{table}[ht]
\centering
\small
\begin{tabular}{ccc}
\toprule
\textbf{mathematician}&\textbf{student}&\textbf{poker player} \\
\midrule
 57.37& 52.45& 47.54\\
\bottomrule
\end{tabular}
\caption{Results on GSM8K. The templates for these roles are shown in Table~\ref{templates} on the Appendix.}
\label{role_immersed}
\end{table}
\paragraph{Effectiveness of NE.} While Nash CoT has demonstrated effectiveness across various LLMs and benchmarks, however, the extent to which Nash CoT's performance is driven by NE remains to be determined. To further investigate, we compare the self-consistency integrated template (denoted as \texttt{self-con.}+\textit{temp} (10 paths)) with Nash CoT (10 paths) using Mistral-Instruct (7B). As shown in Table~\ref{performance_compare_abili}, Nash CoT consistently outperforms \texttt{self-con.}+\textit{temp} when applying the same mathematical templates across selected tasks.
\begin{table}[ht]
\centering
\resizebox{\linewidth}{!}{
\begin{tabular}{c|cccccc}
\toprule
\textbf{Settings}& \textbf{AQuA}	&\textbf{SVAMP}	&\textbf{GSM8K}	&\textbf{AddSub}&\textbf{Avg.} \\
\midrule
\texttt{self-con.}+\textit{temp} &31.7 &71.6 &53.6 &85.8 &60.6 \\
Nash CoT &\textbf{39.9} &\textbf{77.0} &\textbf{55.7} &\textbf{86.3} &\textbf{64.7} \\
\bottomrule
\end{tabular}}
\caption{Ablation of NE. This experiment is conducted to verify the necessity of NE in Nash CoT.}
\label{performance_compare_abili}
\vspace{-10pt}
\end{table}
\section{Conclusion}
We prove the existence of a NE in preferences between role-immersed and normal-status LLM generations, and develop a game system within each path of the multi-path CoT to introduce Nash CoT. Experimental results show that Nash CoT can perform equally or even better than self-consistency while requiring fewer paths. 
\section*{Limitations and Future Work} Since Nash CoT relies on predefined templates, its performance will decline in new scenarios due to limited adaptability in template selection. Future efforts will concentrate on automating the balancing of task feedback and template selection.
\section*{Acknowledgement}This work was supported by the National Science and Technology Innovation 2030 - Major Project (Grant No.2022ZD0208800), and NSFC General Program (Grant No.62176215). Meanwhile, we thank the reviewers for their valuable suggestions.
\bibliography{custom}

\begin{thebibliography}{35}
\providecommand{\natexlab}[1]{#1}

\bibitem[{Akrour et~al.(2011)Akrour, Schoenauer, and Sebag}]{Akrour2011PreferenceBasedPL}
Riad Akrour, Marc Schoenauer, and Mich{\`e}le Sebag. 2011.
\newblock \href {https://api.semanticscholar.org/CorpusID:16505586} {Preference-based policy learning}.
\newblock In \emph{ECML/PKDD}.

\bibitem[{Brown et~al.(2020{\natexlab{a}})Brown, Mann, and Ryder}]{NEURIPS2020_1457c0d6}
Tom Brown, Benjamin Mann, and etc. Ryder. 2020{\natexlab{a}}.
\newblock \href {https://proceedings.neurips.cc/paper_files/paper/2020/file/1457c0d6bfcb4967418bfb8ac142f64a-Paper.pdf} {Language models are few-shot learners}.
\newblock In \emph{Advances in Neural Information Processing Systems}, volume~33, pages 1877--1901. Curran Associates, Inc.

\bibitem[{Brown et~al.(2020{\natexlab{b}})Brown, Mann, Ryder, Subbiah, Kaplan, Dhariwal, Neelakantan, Shyam, Sastry, Askell, Agarwal, Herbert-Voss, Krueger, Henighan, Child, Ramesh, Ziegler, Wu, Winter, Hesse, Chen, Sigler, Litwin, Gray, Chess, Clark, Berner, McCandlish, Radford, Sutskever, and Amodei}]{brown2020language}
Tom~B. Brown, Benjamin Mann, Nick Ryder, Melanie Subbiah, Jared Kaplan, Prafulla Dhariwal, Arvind Neelakantan, Pranav Shyam, Girish Sastry, Amanda Askell, Sandhini Agarwal, Ariel Herbert-Voss, Gretchen Krueger, Tom Henighan, Rewon Child, Aditya Ramesh, Daniel~M. Ziegler, Jeffrey Wu, Clemens Winter, Christopher Hesse, Mark Chen, Eric Sigler, Mateusz Litwin, Scott Gray, Benjamin Chess, Jack Clark, Christopher Berner, Sam McCandlish, Alec Radford, Ilya Sutskever, and Dario Amodei. 2020{\natexlab{b}}.
\newblock \href {https://arxiv.org/abs/2005.14165} {Language models are few-shot learners}.
\newblock \emph{Preprint}, arXiv:2005.14165.

\bibitem[{Busa-Fekete et~al.(2013)Busa-Fekete, Sz{\"o}r{\'e}nyi, Weng, Cheng, and H{\"u}llermeier}]{BusaFekete2013PreferencebasedED}
R{\'o}bert~Istvan Busa-Fekete, Bal{\'a}zs Sz{\"o}r{\'e}nyi, Paul Weng, Weiwei Cheng, and Eyke H{\"u}llermeier. 2013.
\newblock \href {https://api.semanticscholar.org/CorpusID:267824} {Preference-based evolutionary direct policy search}.

\bibitem[{Chen et~al.(2024)Chen, Deng, Yuan, Ji, and Gu}]{chen2024selfplayfinetuningconvertsweak}
Zixiang Chen, Yihe Deng, Huizhuo Yuan, Kaixuan Ji, and Quanquan Gu. 2024.
\newblock \href {https://arxiv.org/abs/2401.01335} {Self-play fine-tuning converts weak language models to strong language models}.
\newblock \emph{Preprint}, arXiv:2401.01335.

\bibitem[{Cheng et~al.(2011)Cheng, F{\"u}rnkranz, H{\"u}llermeier, and Park}]{Cheng2011PreferenceBasedPI}
Weiwei Cheng, Johannes F{\"u}rnkranz, Eyke H{\"u}llermeier, and Sang-Hyeun Park. 2011.
\newblock \href {https://api.semanticscholar.org/CorpusID:17361173} {Preference-based policy iteration: Leveraging preference learning for reinforcement learning}.
\newblock In \emph{ECML/PKDD}.

\bibitem[{Cobbe et~al.(2021)Cobbe, Kosaraju, Bavarian, Chen, Jun, Kaiser, Plappert, Tworek, Hilton, Nakano, Hesse, and Schulman}]{cobbe2021training}
Karl Cobbe, Vineet Kosaraju, Mohammad Bavarian, Mark Chen, Heewoo Jun, Lukasz Kaiser, Matthias Plappert, Jerry Tworek, Jacob Hilton, Reiichiro Nakano, Christopher Hesse, and John Schulman. 2021.
\newblock \href {https://arxiv.org/abs/2110.14168} {Training verifiers to solve math word problems}.
\newblock \emph{Preprint}, arXiv:2110.14168.

\bibitem[{Du et~al.(2022)Du, Qian, Liu, Ding, Qiu, Yang, and Tang}]{du2022glm}
Zhengxiao Du, Yujie Qian, Xiao Liu, Ming Ding, Jiezhong Qiu, Zhilin Yang, and Jie Tang. 2022.
\newblock Glm: General language model pretraining with autoregressive blank infilling.
\newblock In \emph{Proceedings of the 60th Annual Meeting of the Association for Computational Linguistics (Volume 1: Long Papers)}, pages 320--335.

\bibitem[{etc.(2023)}]{touvron2023llama}
Hugo~Touvron etc. 2023.
\newblock \href {https://arxiv.org/abs/2307.09288} {Llama 2: Open foundation and fine-tuned chat models}.
\newblock \emph{Preprint}, arXiv:2307.09288.

\bibitem[{Geva et~al.(2021)Geva, Khashabi, Segal, Khot, Roth, and Berant}]{geva2021did}
Mor Geva, Daniel Khashabi, Elad Segal, Tushar Khot, Dan Roth, and Jonathan Berant. 2021.
\newblock \href {https://arxiv.org/abs/2101.02235} {Did aristotle use a laptop? a question answering benchmark with implicit reasoning strategies}.
\newblock \emph{Preprint}, arXiv:2101.02235.

\bibitem[{GLM et~al.(2024)GLM, Zeng, Xu, Wang, Zhang, Yin, Rojas, Feng, Zhao, Lai, Yu, Wang, Sun, Zhang, Cheng, Gui, Tang, Zhang, Li, Zhao, Wu, Zhong, Liu, Huang, Zhang, Zheng, Lu, Duan, Zhang, Cao, Yang, Tam, Zhao, Liu, Xia, Zhang, Gu, Lv, Liu, Liu, Yang, Song, Zhang, An, Xu, Niu, Yang, Li, Bai, Dong, Qi, Wang, Yang, Du, Hou, and Wang}]{glm2024chatglm}
Team GLM, Aohan Zeng, Bin Xu, Bowen Wang, Chenhui Zhang, Da~Yin, Diego Rojas, Guanyu Feng, Hanlin Zhao, Hanyu Lai, Hao Yu, Hongning Wang, Jiadai Sun, Jiajie Zhang, Jiale Cheng, Jiayi Gui, Jie Tang, Jing Zhang, Juanzi Li, Lei Zhao, Lindong Wu, Lucen Zhong, Mingdao Liu, Minlie Huang, Peng Zhang, Qinkai Zheng, Rui Lu, Shuaiqi Duan, Shudan Zhang, Shulin Cao, Shuxun Yang, Weng~Lam Tam, Wenyi Zhao, Xiao Liu, Xiao Xia, Xiaohan Zhang, Xiaotao Gu, Xin Lv, Xinghan Liu, Xinyi Liu, Xinyue Yang, Xixuan Song, Xunkai Zhang, Yifan An, Yifan Xu, Yilin Niu, Yuantao Yang, Yueyan Li, Yushi Bai, Yuxiao Dong, Zehan Qi, Zhaoyu Wang, Zhen Yang, Zhengxiao Du, Zhenyu Hou, and Zihan Wang. 2024.
\newblock \href {https://arxiv.org/abs/2406.12793} {Chatglm: A family of large language models from glm-130b to glm-4 all tools}.
\newblock \emph{Preprint}, arXiv:2406.12793.

\bibitem[{Hendrycks et~al.(2021)Hendrycks, Burns, Kadavath, Arora, Basart, Tang, Song, and Steinhardt}]{hendrycksmath2021}
Dan Hendrycks, Collin Burns, Saurav Kadavath, Akul Arora, Steven Basart, Eric Tang, Dawn Song, and Jacob Steinhardt. 2021.
\newblock Measuring mathematical problem solving with the math dataset.
\newblock \emph{NeurIPS}.

\bibitem[{Hosseini et~al.(2014)Hosseini, Hajishirzi, Etzioni, and Kushman}]{hosseini-etal-2014-learning}
Mohammad~Javad Hosseini, Hannaneh Hajishirzi, Oren Etzioni, and Nate Kushman. 2014.
\newblock \href {https://doi.org/10.3115/v1/D14-1058} {Learning to solve arithmetic word problems with verb categorization}.
\newblock In \emph{Proceedings of the 2014 Conference on Empirical Methods in Natural Language Processing ({EMNLP})}, pages 523--533, Doha, Qatar. Association for Computational Linguistics.

\bibitem[{Huang et~al.(2022)Huang, Gu, Hou, Wu, Wang, Yu, and Han}]{huang2022large}
Jiaxin Huang, Shixiang~Shane Gu, Le~Hou, Yuexin Wu, Xuezhi Wang, Hongkun Yu, and Jiawei Han. 2022.
\newblock \href {https://arxiv.org/abs/2210.11610} {Large language models can self-improve}.
\newblock \emph{Preprint}, arXiv:2210.11610.

\bibitem[{Jiang et~al.(2023)Jiang, Sablayrolles, Mensch, Bamford, Chaplot, de~las Casas, Bressand, Lengyel, Lample, Saulnier, Lavaud, Lachaux, Stock, Scao, Lavril, Wang, Lacroix, and Sayed}]{jiang2023mistral}
Albert~Q. Jiang, Alexandre Sablayrolles, Arthur Mensch, Chris Bamford, Devendra~Singh Chaplot, Diego de~las Casas, Florian Bressand, Gianna Lengyel, Guillaume Lample, Lucile Saulnier, Lélio~Renard Lavaud, Marie-Anne Lachaux, Pierre Stock, Teven~Le Scao, Thibaut Lavril, Thomas Wang, Timothée Lacroix, and William~El Sayed. 2023.
\newblock \href {https://arxiv.org/abs/2310.06825} {Mistral 7b}.
\newblock \emph{Preprint}, arXiv:2310.06825.

\bibitem[{Jin and Lu(2023)}]{jin2023tabcot}
Ziqi Jin and Wei Lu. 2023.
\newblock \href {https://arxiv.org/abs/2305.17812} {Tab-cot: Zero-shot tabular chain of thought}.
\newblock \emph{Preprint}, arXiv:2305.17812.

\bibitem[{Kojima et~al.(2023)Kojima, Gu, Reid, Matsuo, and Iwasawa}]{kojima2023large}
Takeshi Kojima, Shixiang~Shane Gu, Machel Reid, Yutaka Matsuo, and Yusuke Iwasawa. 2023.
\newblock \href {https://arxiv.org/abs/2205.11916} {Large language models are zero-shot reasoners}.
\newblock \emph{Preprint}, arXiv:2205.11916.

\bibitem[{Koncel-Kedziorski et~al.(2015)Koncel-Kedziorski, Hajishirzi, Sabharwal, Etzioni, and Ang}]{koncel-kedziorski-etal-2015-parsing}
Rik Koncel-Kedziorski, Hannaneh Hajishirzi, Ashish Sabharwal, Oren Etzioni, and Siena~Dumas Ang. 2015.
\newblock \href {https://doi.org/10.1162/tacl_a_00160} {Parsing algebraic word problems into equations}.
\newblock \emph{Transactions of the Association for Computational Linguistics}, 3:585--597.

\bibitem[{Ling et~al.(2017)Ling, Yogatama, Dyer, and Blunsom}]{ling2017program}
Wang Ling, Dani Yogatama, Chris Dyer, and Phil Blunsom. 2017.
\newblock \href {https://arxiv.org/abs/1705.04146} {Program induction by rationale generation : Learning to solve and explain algebraic word problems}.
\newblock \emph{Preprint}, arXiv:1705.04146.

\bibitem[{Munos et~al.(2023)Munos, Valko, Calandriello, Azar, Rowland, Guo, Tang, Geist, Mesnard, Michi, Selvi, Girgin, Momchev, Bachem, Mankowitz, Precup, and Piot}]{munos2023nash}
Rémi Munos, Michal Valko, Daniele Calandriello, Mohammad~Gheshlaghi Azar, Mark Rowland, Zhaohan~Daniel Guo, Yunhao Tang, Matthieu Geist, Thomas Mesnard, Andrea Michi, Marco Selvi, Sertan Girgin, Nikola Momchev, Olivier Bachem, Daniel~J. Mankowitz, Doina Precup, and Bilal Piot. 2023.
\newblock \href {https://arxiv.org/abs/2312.00886} {Nash learning from human feedback}.
\newblock \emph{Preprint}, arXiv:2312.00886.

\bibitem[{OpenAI(2024)}]{openai2024gpt4}
OpenAI. 2024.
\newblock \href {https://arxiv.org/abs/2303.08774} {Gpt-4 technical report}.
\newblock \emph{Preprint}, arXiv:2303.08774.

\bibitem[{Ouyang et~al.(2022)Ouyang, Wu, Jiang, Almeida, Wainwright, Mishkin, Zhang, Agarwal, Slama, Ray, Schulman, Hilton, Kelton, Miller, Simens, Askell, Welinder, Christiano, Leike, and Lowe}]{ouyang2022training}
Long Ouyang, Jeff Wu, Xu~Jiang, Diogo Almeida, Carroll~L. Wainwright, Pamela Mishkin, Chong Zhang, Sandhini Agarwal, Katarina Slama, Alex Ray, John Schulman, Jacob Hilton, Fraser Kelton, Luke Miller, Maddie Simens, Amanda Askell, Peter Welinder, Paul Christiano, Jan Leike, and Ryan Lowe. 2022.
\newblock \href {https://arxiv.org/abs/2203.02155} {Training language models to follow instructions with human feedback}.
\newblock \emph{Preprint}, arXiv:2203.02155.

\bibitem[{Patel et~al.(2021)Patel, Bhattamishra, and Goyal}]{patel-etal-2021-nlp}
Arkil Patel, Satwik Bhattamishra, and Navin Goyal. 2021.
\newblock \href {https://doi.org/10.18653/v1/2021.naacl-main.168} {Are {NLP} models really able to solve simple math word problems?}
\newblock In \emph{Proceedings of the 2021 Conference of the North American Chapter of the Association for Computational Linguistics: Human Language Technologies}, pages 2080--2094, Online. Association for Computational Linguistics.

\bibitem[{Radford et~al.(2019)Radford, Wu, Child, Luan, Amodei, and Sutskever}]{Radford2019LanguageMA}
Alec Radford, Jeff Wu, Rewon Child, David Luan, Dario Amodei, and Ilya Sutskever. 2019.
\newblock \href {https://api.semanticscholar.org/CorpusID:160025533} {Language models are unsupervised multitask learners}.

\bibitem[{Rae et~al.(2022)Rae, Borgeaud, Cai, Millican, Hoffmann, Song, Aslanides, Henderson, Ring, Young, Rutherford, Hennigan, Menick, Cassirer, Powell, van~den Driessche, Hendricks, Rauh, Huang, Glaese, Welbl, Dathathri, Huang, Uesato, Mellor, Higgins, Creswell, McAleese, Wu, Elsen, Jayakumar, Buchatskaya, Budden, Sutherland, Simonyan, Paganini, Sifre, Martens, Li, Kuncoro, Nematzadeh, Gribovskaya, Donato, Lazaridou, Mensch, Lespiau, Tsimpoukelli, Grigorev, Fritz, Sottiaux, Pajarskas, Pohlen, Gong, Toyama, de~Masson~d'Autume, Li, Terzi, Mikulik, Babuschkin, Clark, de~Las~Casas, Guy, Jones, Bradbury, Johnson, Hechtman, Weidinger, Gabriel, Isaac, Lockhart, Osindero, Rimell, Dyer, Vinyals, Ayoub, Stanway, Bennett, Hassabis, Kavukcuoglu, and Irving}]{rae2022scaling}
Jack~W. Rae, Sebastian Borgeaud, Trevor Cai, Katie Millican, Jordan Hoffmann, Francis Song, John Aslanides, Sarah Henderson, Roman Ring, Susannah Young, Eliza Rutherford, Tom Hennigan, Jacob Menick, Albin Cassirer, Richard Powell, George van~den Driessche, Lisa~Anne Hendricks, Maribeth Rauh, Po-Sen Huang, Amelia Glaese, Johannes Welbl, Sumanth Dathathri, Saffron Huang, Jonathan Uesato, John Mellor, Irina Higgins, Antonia Creswell, Nat McAleese, Amy Wu, Erich Elsen, Siddhant Jayakumar, Elena Buchatskaya, David Budden, Esme Sutherland, Karen Simonyan, Michela Paganini, Laurent Sifre, Lena Martens, Xiang~Lorraine Li, Adhiguna Kuncoro, Aida Nematzadeh, Elena Gribovskaya, Domenic Donato, Angeliki Lazaridou, Arthur Mensch, Jean-Baptiste Lespiau, Maria Tsimpoukelli, Nikolai Grigorev, Doug Fritz, Thibault Sottiaux, Mantas Pajarskas, Toby Pohlen, Zhitao Gong, Daniel Toyama, Cyprien de~Masson~d'Autume, Yujia Li, Tayfun Terzi, Vladimir Mikulik, Igor Babuschkin, Aidan Clark, Diego de~Las~Casas, Aurelia Guy, Chris Jones,
  James Bradbury, Matthew Johnson, Blake Hechtman, Laura Weidinger, Iason Gabriel, William Isaac, Ed~Lockhart, Simon Osindero, Laura Rimell, Chris Dyer, Oriol Vinyals, Kareem Ayoub, Jeff Stanway, Lorrayne Bennett, Demis Hassabis, Koray Kavukcuoglu, and Geoffrey Irving. 2022.
\newblock \href {https://arxiv.org/abs/2112.11446} {Scaling language models: Methods, analysis \& insights from training gopher}.
\newblock \emph{Preprint}, arXiv:2112.11446.

\bibitem[{Rafailov et~al.(2023)Rafailov, Sharma, Mitchell, Ermon, Manning, and Finn}]{rafailov2023direct}
Rafael Rafailov, Archit Sharma, Eric Mitchell, Stefano Ermon, Christopher~D. Manning, and Chelsea Finn. 2023.
\newblock \href {https://arxiv.org/abs/2305.18290} {Direct preference optimization: Your language model is secretly a reward model}.
\newblock \emph{Preprint}, arXiv:2305.18290.

\bibitem[{Rosen(1964)}]{1964Existence}
J.~B Rosen. 1964.
\newblock Existence and uniqueness of equilibrium points for concave n-person games.

\bibitem[{Roy and Roth(2016)}]{roy2016solving}
Subhro Roy and Dan Roth. 2016.
\newblock \href {https://arxiv.org/abs/1608.01413} {Solving general arithmetic word problems}.
\newblock \emph{Preprint}, arXiv:1608.01413.

\bibitem[{Sion(1958)}]{Sion1958OnGM}
Maurice Sion. 1958.
\newblock \href {https://api.semanticscholar.org/CorpusID:120295759} {On general minimax theorems}.
\newblock \emph{Pacific Journal of Mathematics}, 8:171--176.

\bibitem[{Talmor et~al.(2019)Talmor, Herzig, Lourie, and Berant}]{talmor2019commonsenseqa}
Alon Talmor, Jonathan Herzig, Nicholas Lourie, and Jonathan Berant. 2019.
\newblock \href {https://arxiv.org/abs/1811.00937} {Commonsenseqa: A question answering challenge targeting commonsense knowledge}.
\newblock \emph{Preprint}, arXiv:1811.00937.

\bibitem[{Wang et~al.(2023)Wang, Wei, Schuurmans, Le, Chi, Narang, Chowdhery, and Zhou}]{wang2023selfconsistency}
Xuezhi Wang, Jason Wei, Dale Schuurmans, Quoc Le, Ed~Chi, Sharan Narang, Aakanksha Chowdhery, and Denny Zhou. 2023.
\newblock \href {https://arxiv.org/abs/2203.11171} {Self-consistency improves chain of thought reasoning in language models}.
\newblock \emph{Preprint}, arXiv:2203.11171.

\bibitem[{Wei et~al.(2023)Wei, Wang, Schuurmans, Bosma, Ichter, Xia, Chi, Le, and Zhou}]{wei2023chainofthought}
Jason Wei, Xuezhi Wang, Dale Schuurmans, Maarten Bosma, Brian Ichter, Fei Xia, Ed~Chi, Quoc Le, and Denny Zhou. 2023.
\newblock \href {https://arxiv.org/abs/2201.11903} {Chain-of-thought prompting elicits reasoning in large language models}.
\newblock \emph{Preprint}, arXiv:2201.11903.

\bibitem[{Wilson et~al.(2012)Wilson, Fern, and Tadepalli}]{Wilson2012ABA}
Aaron Wilson, Alan Fern, and Prasad Tadepalli. 2012.
\newblock \href {https://api.semanticscholar.org/CorpusID:6019958} {A bayesian approach for policy learning from trajectory preference queries}.
\newblock In \emph{Neural Information Processing Systems}.

\bibitem[{Zeng et~al.(2022)Zeng, Liu, Du, Wang, Lai, Ding, Yang, Xu, Zheng, Xia et~al.}]{zeng2022glm}
Aohan Zeng, Xiao Liu, Zhengxiao Du, Zihan Wang, Hanyu Lai, Ming Ding, Zhuoyi Yang, Yifan Xu, Wendi Zheng, Xiao Xia, et~al. 2022.
\newblock Glm-130b: An open bilingual pre-trained model.
\newblock \emph{arXiv preprint arXiv:2210.02414}.

\bibitem[{Zhang et~al.(2022)Zhang, Zhang, Li, and Smola}]{zhang2022automatic}
Zhuosheng Zhang, Aston Zhang, Mu~Li, and Alex Smola. 2022.
\newblock \href {https://arxiv.org/abs/2210.03493} {Automatic chain of thought prompting in large language models}.
\newblock \emph{Preprint}, arXiv:2210.03493.

\end{thebibliography}
\appendix
\onecolumn
\newpage
\section*{Ethics Claims}
Pre-training or fine-tuning an LLM requires much more computing resources. Multi-path CoT is an ideal approach that has been proven to enhance the performance of LLMs' inference. Meanwhile, Nash CoT effectively reduces the inference paths needed and times of multi-path inference. We believe our approach can further elevate the effectiveness of multi-path inference, thereby further improving the effectiveness of LLM. 
\section{Dataset\label{appendix_data}}
 Our majority dataset are composed of three different kinds of inference tasks. 

\begin{itemize}
\item \textit{Arabic Reasoning:} SingleEq~\citep{koncel-kedziorski-etal-2015-parsing}, AddSub~\citep{hosseini-etal-2014-learning}, MultiArith~\citep{roy2016solving}, GSM8K~\citep{cobbe2021training}, AQUA~\citep{ling2017program}, SVAMP~\citep{patel-etal-2021-nlp}, and MATH dataset~\citep{hendrycksmath2021}. In particular, MATH dataset are process by our own, : When constructing the dataset, we identify the answers to each question in the MATH dataset through ground truth (GT) reasoning, retaining a certain number of questions for each category. When testing the LLM, we input the questions, collect the responses from the deployed LLM, and determine whether the LLM's reasoning is correct by comparing whether the inferred results match the GT, examples are shown in Table~\ref{cases_MATH}. Meanwhile, the dataset or its url will be released at our codebase page.

\begin{table}[ht]
\centering
\begin{tabular}{c|c}
\toprule
Dataset& Capacity  \\
\midrule
Algebra& 778 \\
Counting and Probability& 291\\
Geometry& 235\\
Intermediate Algebra& 443\\
Number Theory& 474\\
Prealgebra& 618\\
Precalculus& 133\\
\bottomrule
\end{tabular}
\caption{Data type and data capacity}
\label{tab:my_label}
\end{table}

\item \textit{Symbolic Reasoning:} Last Letters, Coin Flip~\citep{wei2023chainofthought}, and Object Tracking, Bigbench Date. 
 
\item  \textit{Commonsense Question Answering:} CommonsenseQA~\cite{talmor2019commonsenseqa} and StrategyQA~\cite{geva2021did}. For more details about the dataset please refer to~\cite{wang2023selfconsistency}.
\end{itemize}

\section{Experimental Settings}
In this section, we compare Nash CoT with self-consistency, focusing specifically on self-consistency as a representative method. As illustrated in Figure~\ref{nash_cot}, Nash CoT operates with two distinct loops, where the final answer is derived from voting on the response that achieves Preference Equilibrium most frequently. In contrast, self-consistency determines the most frequent answer across all predictions. 
 \begin{figure*}[ht]
\centering\hspace{-10pt}
\includegraphics[scale=0.45]{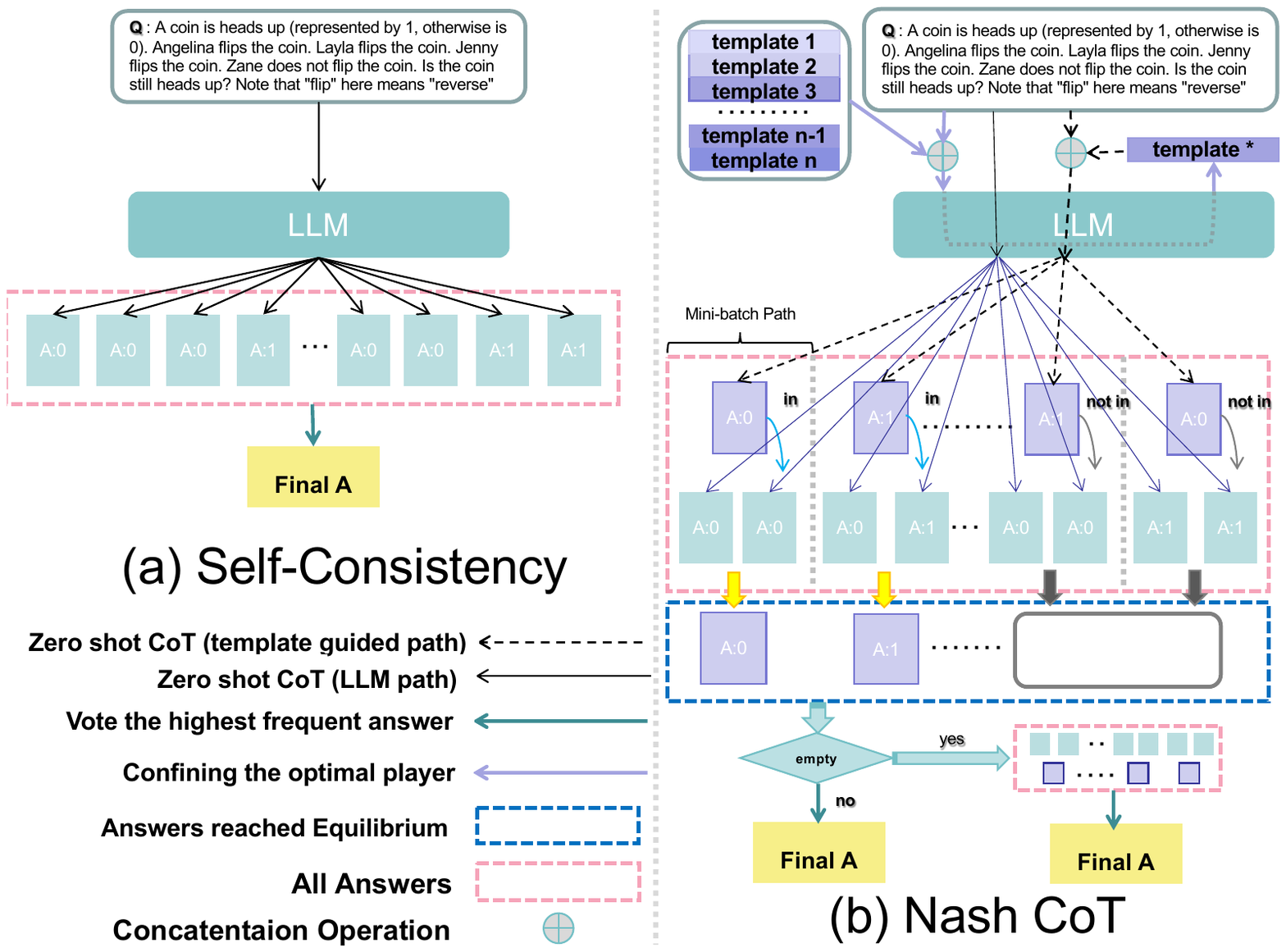}
\vspace{-10pt}
\caption{Comparison of Nash CoT and self-consistency. (right) Nash Chain of thought (Nash CoT). (left) self-consistency. }
\label{nash_cot_compare}
\vspace{-2mm}
\end{figure*}

\section{Usage of LLM.} We utilize LLM to rectify grammar errors.

\section{Computing Resources} Our experiments were run on a computer cluster with 32GB RAM, 4-Core CPU, and NVIDIA-A100 (80G, 32G)/NVIDIA-V100 (32G) GPU, Linux platform.

\section{Source Code.} We have provided source code for reference. Additionally, our code are based on~\url{https://github.com/amazon-science/auto-cot} and refer to the coding manner from~\url{https://github.com/eureka-research/Eureka}.
\newpage
\section{Proof of Theorem~\ref{lemma1}.\label{proof}}

Our proof is refereed to~\citeauthor{munos2023nash}

Here, we examine a preference model incorporating a Kullback-Leibler (KL) divergence constraint. Unlike the trainable system proposed by~\citeauthor{munos2023nash}, which aims to serve as the policy generating the training data for the preference model or address scenarios requiring adherence to a known safe policy, our inference system is non-trainable. Consequently, a static distance is maintained between the players $(\pi_1,\pi_2)$ and a static reference policy $\mu$. The introduction of this regularization term is intended to facilitate a theoretical analysis demonstrating that, under the KL constraint, the game system possesses a unique Nash Equilibrium (NE).

Subsequently, we follow~\citeauthor{munos2023nash} to add a KL regularization term the the preference optimization objective $\ie$

\begin{align}
\begin{split}
\mathcal{P}_{\tau}(\pi_1\prec\pi_2)\overset{def}{=}\mathcal{P}(\pi_1\prec\pi_2)-\tau\cdot {\rm KL}(\pi_2(\cdot|x)||\mu(\cdot|x))+\tau\cdot {\rm KL}(\pi_1(\cdot|x)||\mu(\cdot|x)).
\end{split}
\label{regularized}
\end{align}

Then we prove the existence of NE. Specifically, the mappings $\pi_1\mapsto \mathcal{P}(\pi_1\prec \pi_2)$ and $\pi_2\mapsto \mathcal{P}(\pi_1\prec \pi_2)$ are linear in $\pi_1$ (respectively in $\pi_2$) thus $\pi_1\mapsto \mathcal{P}_{\tau}(\pi_1\prec \pi_2)$ is concave
and $\pi_2\mapsto \mathcal{P}_{\tau}(\pi_1\prec \pi_2)$ is convex. Existence of a NE is derived from the mini-max theorem for convex-concave functions~\cite{Sion1958OnGM}.

Next, we substitute Equation~\ref{regularized} into our problem setup. In our setup, model training is not involved, so $\mu$ can be regarded as one of the strategies in $\pi_1$ and $\pi_2$. Furthermore, we assume $\mu$ to be $\pi_1$, and $\pi_2$ is the role immersed policy, so we can rewrite the Equation~\ref{regularized} as a two-player game system, with the pay-off of $\pi_1$ and $\pi_2$:

\begin{align}
    \begin{split}
        R(\pi_1;\pi_2)&= \mathcal{P}(\pi_2\prec \pi_1)-\tau\cdot  {\rm KL}(\pi_1||\mu) \\
        &=\mathcal{P}(\pi_2\prec \pi_1)
    \end{split}
\end{align}

and

\begin{align}
    \begin{split}
        R(\pi_2;\pi_1)&= \mathcal{P}(\pi_1\prec \pi_2)-\tau\cdot {\rm KL}(\pi_2||\mu) \\
         &\leq \mathcal{P}(\pi_1\prec \pi_2)
    \end{split}
\end{align}

In particular, when the difference between $\pi_2$ and $\mu$ is small enough, R($\pi_2$; $\pi_1$) can be approximately regarded as $\mathcal{P}(\pi_1 \prec \pi_2)$. When the difference between $\pi_2$ and $\mu$ is large enough, $\mathcal{P}(\pi_1 \prec \pi_2)$ serves as an upper bound for $R(\pi_2; \pi_1)$. Therefore, the KL term plays a constraining term in the role-immersed generation process.

Subsequently, to show this game has a unique NE, we have to show the variational inequality is strictly monotone (\citeauthor{1964Existence}, Theorem 2). Meanwhile, followed by the proof of~\citeauthor{munos2023nash} We learn that every NE of the game should satisfy the variational inequality for all $\bar \pi$:
\begin{align}
    \begin{split}
        v^T(\bar \pi^*)(\bar \pi^*-\bar \pi)\leq 0,
    \end{split}
\end{align}
Furthermore, the inequality is strictly monotone if and only if for every $\bar \pi_1$ and $\bar \pi_2$ we have that:
\begin{align}
    \begin{split}
        (v(\bar \pi_1)-v(\bar \pi_2))^T(\bar \pi_1-\bar \pi_2)\leq 0,
    \end{split}
\end{align}
And this inequality has been proved by~\citeauthor{munos2023nash}, with equality only at $\bar \pi_1=\bar \pi_2$.

Therefore, $\pi_1=\pi_2$
  is an essential condition for maintaining strictly monotone. Meanwhile, strictly monotone serves as the guarantee of the uniqueness of the NE. Therefore, we selected  
$\pi_1=\pi_2$ as an indicator to ensure the uniqueness of the NE.
\section{Extra Explanation}
\paragraph{Why in Table~\ref{tab:main_label1},~\ref{tab:main_label2} and~\ref{tab:my_labe3} do we set the number of pathways for Nash CoT to be half of self-consistency?} If Nash CoT can surpass self-consistency with half the number of reasoning pathways, then it is easier to demonstrate the validity and efficiency of Nash CoT. Because that according to the experimental conclusion in~\citeauthor{wang2023selfconsistency} that \texttt{``The experimental effect of self-consistency increases as the number of pathways increases."}, if Nash-CoT outperforms self-consistency with fewer paths, it indicates that Nash-CoT is better than self-consistency across all settings with an equal or greater number of paths.
\paragraph{Which one performs better between Nash CoT and self-consistency when they have the same number of pathways?} In our ablation experiments to test the impact of the number of pathways on Nash CoT, we found that by adjusting the number of pathways, Nash CoT can surpass self-consistency with fewer pathways on Aqua, CommensenseQA, and Object Tracking. Additionally, in Table~\ref{larger_llm_performance}, we used GLM4 (large), Mistral-Instruct (7B) to outperform self-consistency on the MATH dataset with the same number of pathways.
\newpage
\section{Case Study}
\subsection{LLM as preference model\label{preference_model}}
In this experiment, we provided examples of LLM inferences to demonstrate that LLMs can select the most appropriate template based on the question. Specifically, in Table~\ref{templates}, we set up a series of templates for different roles, arranging them in different sequences and labeling them with numbers. These were input into the model, allowing it to rank the templates according to its preference for the question. We observe in Table~\ref{tab:ques_template} that the order given by the LLM corresponded to the performance strengths of different templates on the respective tasks, which proves that LLMs can directly serve as preference models to select the optimal template for a specific question.
\begin{table*}[ht]
\centering
\small
\begin{tabularx}{\textwidth}{cXXX}
\toprule
\textbf{Player}&\textbf{Description}&\textbf{Correlation}\\
\midrule
Mathematician&	You are a mathematician, You excel at analyzing problems from a Mathematical logical perspective and arrive at conclusions that align with your values.&	High\\
\hline
Student	&You are a student, please answer the following questions.	&Medium\\
\hline
Poker Player&You are playing a poker game, please do your best to play poker games.	&Low\\
\bottomrule
\end{tabularx}
\caption{Player templates we utilized.}
\label{templates}
\end{table*}
\begin{table*}[ht]
\centering
\small
\resizebox{\textwidth}{!}{
\begin{tabularx}{\textwidth}{lXXX}
\toprule
&\textbf{Input} & \textbf{Ranking} \\
\midrule
&We have four templates: 1. Mathematician: You are a mathematician, You excel at analyzing problems from a Mathematical logical perspective and arrive at conclusions that align with your values. 2. Student: You are a student, please answer the following questions. 3. Poker Player: You are playing a poker game, please do your best to play poker games. Please rank their correlations to a math problem, and directly output the sequential ranked numerical tag.& 1, 2, 3 \\
\midrule
&We have four templates: 1. Student: You are a student, please answer the following questions. 2. Mathematician: You are a mathematician, You excel at analyzing problems from a Mathematical logical perspective and arrive at conclusions that align with your values. 3. Poker Player: You are playing poker game, please do your best to play poker games. Please rank their correlations to math problem, and directly output the sequential ranked numerical tag.&2, 1, 3\\
\bottomrule
\end{tabularx}
}
\caption{We provide cases for the ranking provided by LLM. The roles of Mathematician, Student, and Poker Player achieved scores of 57.37, 52.45, and 47.54, respectively, on GSM8K. The performance of these different roles is consistent with the model's ranking.}
\label{tab:ques_template}
\end{table*}
Meanwhile, we utilize ChatGPT to rank these templates based on their relevance to the topic of questions. Our inputs are shown in Table~\ref{tab:ques_template}. The model's ranking of different roles is consistent with their performance levels on the mathematical dataset. Meanwhile, changing the input order of the templates simultaneously does not affect the model's ranking of the templates.
\newpage
\section{Format of our processed MATH dataset}
We provide several cases for readers to understand the format of our processed MATH dataset.
\begin{table*}[ht]
\centering
\small
\resizebox{\textwidth}{!}{
\begin{tabularx}{\textwidth}{lXcc}
\toprule
\textbf{Dataset} &\textbf{Question} & \textbf{Answer}&\textbf{Capacity} \\
\midrule
Algebra&How many vertical asymptotes does the graph of $y=\frac{2}{x^2+x-6}$ have?&2&778\\
\midrule
Counting and Probability&$n$ fair 6-sided dice are simultaneously rolled. The probability that exactly two of them show a number other than 1 is $\frac{25}{216}$. Find $n$.&4&291\\
\midrule
Geometry&We have triangle $\triangle ABC$ where $AB = AC$ and $AD$ is an altitude. Meanwhile, $E$ is a point on $AC$ such that $AB \parallel DE.$ If $BC = 12$  and the area of $\triangle ABC$ is $180,$ what is the area of $ABDE$?&135&235\\
\midrule
Intermediate Algebra&Find the sum of all complex roots of the equation \[\frac{1}{x-1} + \frac{1}{x-5} + \frac{1}{x-10} + \frac{1}{x-25} = 2,\]given that there are no repeated roots.&43&443\\
\midrule
Number Theory&Kirsty needs to hire a plumber to fix her house. The plumber charges $242_5$ dollars for every hour of labor and $367_{8}$ dollars for equipment. If the plumber works for $3.5_{10}$ hours, how many dollars (in base ten) will Kirsty owe the plumber?&499&474\\
\midrule
Prealgebra&John and Gary are playing a game. John spins a spinner numbered with integers from 1 to 20. Gary then writes a list of all of the positive factors of the number spun except for the number itself. Gary then creates a new spinner with all of the numbers on his list. John then spins this spinner, and the process continues. The game is over when the spinner has no numbers on it. If John spins a 20 on his first spin, what is the maximum number of total spins (including the one he already made) that John can make before the game is over?&4&618\\
\midrule
Precalculus&Let $\mathbf{a}$ and $\mathbf{b}$ be vectors such that
\[\mathbf{v} = \operatorname{proj}_{\mathbf{a}} \mathbf{v} + \operatorname{proj}_{\mathbf{b}} \mathbf{v}\]for all vectors $\mathbf{v}.$  Enter all possible values of $\mathbf{a} \cdot \mathbf{b},$ separated by commas.&0&133\\
\bottomrule
\end{tabularx}
}
\caption{Examples of MATH dataset. We give the format of our processed MATH dataset. Meanwhile, the dataset or its URL will be available on our codebase page upon release.}
\label{cases_MATH}
\end{table*}
\end{document}